\documentclass{article}
\pdfoutput=1

\usepackage{times}
\usepackage{graphicx}
\usepackage{subfigure} 
\usepackage{algorithm}
\usepackage{algorithmic}
\usepackage{amsmath}
\usepackage{amsfonts}
\allowdisplaybreaks[0]
\usepackage{enumitem}
\usepackage[super]{nth}

\PassOptionsToPackage{hyphens}{url}\usepackage{hyperref}

\usepackage[accepted]{icml2015} 

\renewcommand{\vec}[1]{\mathbf{#1}}
\def\equationautorefname~#1\null{%
  Equation~(#1)\null
}

\begin{document}

\twocolumn[
\icmltitle{Highway Networks}

\icmlauthor{Rupesh Kumar Srivastava}{rupesh@idsia.ch}
\icmlauthor{Klaus Greff}{klaus@idsia.ch}
\icmlauthor{J{\"u}rgen Schmidhuber}{juergen@idsia.ch}
\icmladdress{\begin{center}%
The Swiss AI Lab IDSIA\\
Istituto Dalle Molle di Studi sull'Intelligenza Artificiale\\
Universit\`{a} della Svizzera italiana (USI)\\
Scuola universitaria professionale della Svizzera italiana (SUPSI)\\
Galleria 2, 6928 Manno-Lugano, Switzerland\end{center}
}

\icmlkeywords{deep learning}

\vskip 0.3in
]

\begin{abstract}
There is plenty of theoretical and empirical evidence that depth of neural networks is a crucial ingredient for their success. However, network training becomes more difficult with increasing depth and training of very deep networks remains an open problem.
In this extended abstract, we introduce a new architecture designed to ease gradient-based training of very deep networks. 
We refer to networks with this architecture as highway networks, since they allow unimpeded information flow across several layers on \emph{information highways}. 
The architecture is characterized by the use of gating units which learn to regulate the flow of information through a network.
Highway networks with hundreds of layers can be trained directly using stochastic gradient descent and with a variety of activation functions, opening up the possibility of studying extremely deep and efficient architectures.

\end{abstract}

\textbf{Note}: A full paper extending this study is available at \url{http://arxiv.org/abs/1507.06228}, with additional references, experiments and analysis.

\section{Introduction}

Many recent empirical breakthroughs in supervised machine learning have been achieved through the application of deep neural networks. Network depth (referring to the number of successive computation layers) has played perhaps the most important role in these successes. For instance, the top-5 image classification accuracy on the 1000-class ImageNet dataset has increased from $\sim$84\% \citep{Krizhevsky2012} to $\sim$95\% \citep{Szegedy2014,Simonyan2014} through the use of ensembles of deeper architectures and smaller receptive fields \citep{Ciresan2011a,Ciresan2011,Ciresan2012} in just a few years. 

On the theoretical side, it is well known that deep networks can represent certain function classes exponentially more efficiently than shallow ones (e.g. the work of \citet{Hastad1987,Hastad1991} and recently of \citet{Montufar2014}). As argued by \citet{Bengio2013}, the use of deep networks can offer both computational and statistical efficiency for complex tasks.

However, training deeper networks is not as straightforward as simply adding layers. Optimization of deep networks has proven to be considerably more difficult, leading to research on initialization schemes \citep{Glorot2010,Saxe2013,He2015}, techniques of training networks in multiple stages \citep{Simonyan2014,Romero2014} or with temporary companion loss functions attached to some of the layers \citep{Szegedy2014,Lee2015}. 

In this extended abstract, we present a novel architecture that enables the optimization of networks with virtually arbitrary depth.
This is accomplished through the use of a learned gating mechanism for regulating information flow which is inspired by Long Short Term Memory recurrent neural networks \cite{Hochreiter1995}.
Due to this gating mechanism, a neural network can have paths along which information can flow across several layers without attenuation. We call such paths \emph{information highways}, and such networks \emph{highway networks}. 

In preliminary experiments, we found that highway networks as deep as 900 layers can be optimized using simple Stochastic Gradient Descent (SGD) with momentum. For up to 100 layers we compare their training behavior to that of traditional networks with normalized initialization \citep{Glorot2010,He2015}. We show that optimization of highway networks is virtually independent of depth, while for traditional networks it suffers significantly as the number of layers increases. We also show that architectures comparable to those recently presented by \citet{Romero2014} can be directly trained to obtain similar test set accuracy on the  CIFAR-10 dataset without the need for a pre-trained teacher network.

\subsection{Notation}
We use boldface letters for vectors and matrices, and italicized capital letters to denote transformation functions. $\mathbf{0}$ and $\mathbf{1}$ denote vectors of zeros and ones respectively, and $\vec{I}$ denotes an identity matrix. The function $\sigma(x)$ is defined as $\sigma(x) = \frac{1}{1+e^{-x}}, x \in \mathbb{R}$.

\section{Highway Networks}

A \emph{plain} feedforward neural network typically consists of $L$ layers where the $l^{th}$ layer ($l \in \{1, 2, ..., L\}$) applies a non-linear transform $H$ (parameterized by $\vec{W_{H,l}}$) on its input $\vec{x_l}$ to produce its output $\vec{y_l}$. Thus, $\vec{x_1}$ is the input to the network and $\vec{y_L}$ is the network's output. Omitting the layer index and biases for clarity,

\begin{equation}\label{eq:plain}
\vec{y} = H(\vec{x}, \vec{W_H}).
\end{equation}

$H$ is usually an affine transform followed by a non-linear activation function, but in general it may take other forms.

For a highway network, we additionally define two non-linear transforms $T(\vec{x}, \vec{W_T})$ and $C(\vec{x}, \vec{W_C})$ such that

\begin{equation}\label{eq:highway}
\vec{y} = H(\vec{x}, \vec{W_H}) \cdotp T(\vec{x}, \vec{W_T}) + \vec{x} \cdot C(\vec{x}, \vec{W_C}).
\end{equation}

We refer to $T$ as the \emph{transform} gate and $C$ as the \emph{carry} gate, since they express how much of the output is produced by transforming the input and carrying it, respectively. For simplicity, in this paper we set $C = 1 - T$, giving

\begin{equation}\label{eq:highway-simple}
\vec{y} = H(\vec{x}, \vec{W_H}) \cdotp T(\vec{x}, \vec{W_T}) + \vec{x} \cdot (1 - T(\vec{x}, \vec{W_T})).
\end{equation}

The dimensionality of $\vec{x}, \vec{y}, H(\vec{x}, \vec{W_H})$ and $T(\vec{x}, \vec{W_T})$ must be the same for \autoref{eq:highway-simple} to be valid.
Note that this re-parametrization of the layer transformation is much more flexible than \autoref{eq:plain}. 
In particular, observe that 

\begin{equation}\label{eq:highway-conditions}
\vec{y} = 
\begin{cases}
	\vec{x}, &\text{if } T(\vec{x}, \vec{W_T}) = \mathbf{0},\\
	H(\vec{x}, \vec{W_H}), &\text{if } T(\vec{x}, \vec{W_T}) = \mathbf{1}.
\end{cases}
\end{equation}

Similarly, for the Jacobian of the layer transform,

\begin{equation}
\frac{d\vec{y}}{d\vec{x}} = 
\begin{cases}
	\vec{I}, &\text{if } T(\vec{x}, \vec{W_T}) = \mathbf{0},\\
	H'(\vec{x}, \vec{W_H}), &\text{if } T(\vec{x}, \vec{W_T}) = \mathbf{1}.
\end{cases}
\end{equation}

Thus, depending on the output of the transform gates, a highway layer can smoothly vary its behavior between that of a plain layer and that of a layer which simply passes its inputs through. Just as a plain layer consists of multiple computing units such that the $i^{th}$ unit computes $y_i = H_i(\vec{x})$, a highway network consists of multiple blocks such that the $i^{th}$ block computes a \emph{block state} $H_i(\vec{x})$ and \emph{transform gate output} $T_i(\vec{x})$. Finally, it produces the \emph{block output} $y_i = H_i(\vec{x})*T_i(\vec{x}) + x_i*(1 - T_i(\vec{x}))$, which is connected to the next layer. 

\subsection{Constructing Highway Networks}

As mentioned earlier, \autoref{eq:highway-simple} requires that the dimensionality of $\vec{x}, \vec{y}, H(\vec{x}, \vec{W_H})$ and $T(\vec{x}, \vec{W_T})$ be the same. In cases when it is desirable to change the size of the representation, one can replace $\vec{x}$ with $\vec{\hat{x}}$ obtained by suitably sub-sampling or zero-padding $\vec{x}$. Another alternative is to use a plain layer (without highways) to change dimensionality and then continue with stacking highway layers. This is the alternative we use in this study.

Convolutional highway layers are constructed similar to fully connected layers. Weight-sharing and local receptive fields are utilized for both $H$ and $T$ transforms. We use zero-padding to ensure that the block state and transform gate feature maps are the same size as the input.

\subsection{Training Deep Highway Networks}

For plain deep networks, training with SGD stalls at the beginning unless a specific weight initialization scheme is used such that the variance of the signals during forward and backward propagation is preserved initially \citep{Glorot2010,He2015}. This initialization depends on the exact functional form of $H$.

For highway layers, we use the transform gate defined as $T(\vec{x}) = \sigma(\vec{W_T}^T\vec{x} + \vec{b_T})$, where $\vec{W_T}$ is the weight matrix and $\vec{b_T}$ the bias vector for the transform gates. This suggests a simple initialization scheme which is independent of the nature of $H$: $b_T$ can be initialized with a negative value (e.g. -1, -3 etc.) such that the network is initially biased towards \emph{carry} behavior. This scheme is strongly inspired by the proposal of \citet{Gers1999} to initially bias the gates in a Long Short-Term Memory recurrent network to help bridge long-term temporal dependencies early in learning. Note that $\sigma(x) \in (0, 1), \forall x \in \mathbb{R}$, so the conditions in \autoref{eq:highway-conditions} can never be exactly true. 

In our experiments, we found that a negative bias initialization was sufficient for learning to proceed in very deep networks for various zero-mean initial distributions of $W_H$ and different activation functions used by $H$. This is significant property since in general it may not be possible to find effective initialization schemes for many choices of $H$.

\begin{figure*}[t]
\includegraphics[width=\textwidth]{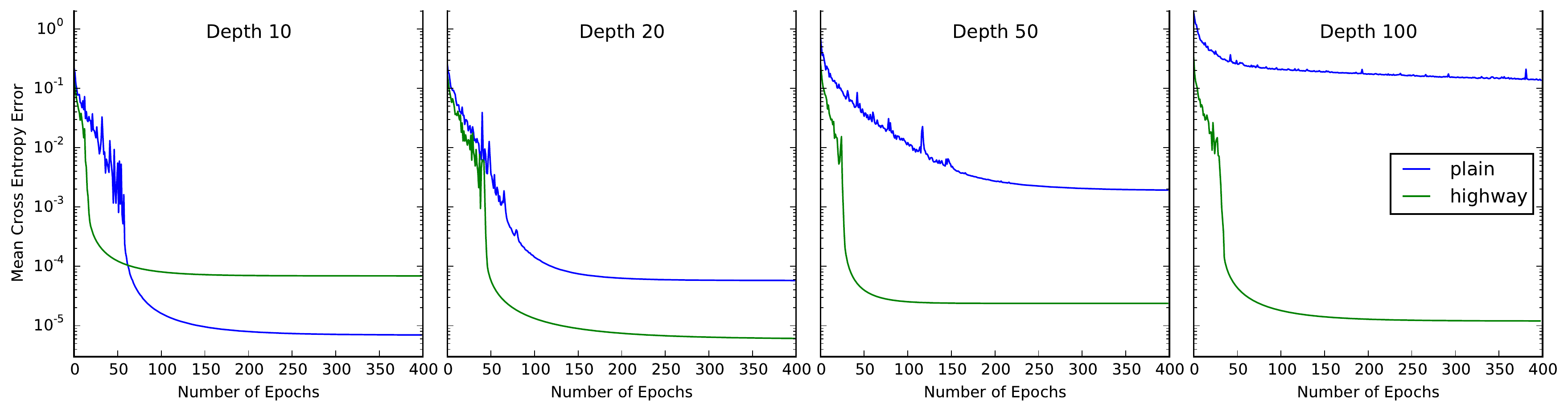}
\caption{Comparison of optimization of plain networks and highway networks of various depths. All networks were optimized using SGD with momentum. The curves shown are for the best hyperparameter settings obtained for each configuration using a random search. Plain networks become much harder to optimize with increasing depth, while highway networks with up to 100 layers can still be optimized well.}
\label{fig:mnist-convergence}
\end{figure*}

\section{Experiments}

\subsection{Optimization}
Very deep plain networks become difficult to optimize even if using the variance-preserving initialization scheme form \cite{He2015}. 
To show that highway networks do not suffer from depth in the same way we train run a series of experiments on the MNIST digit classification dataset. 
We measure the cross entropy error on the training set, to investigate optimization, without conflating them with generalization issues.

We train both plain networks and highway networks with the same architecture and varying depth. The first layer is always a regular fully-connected layer followed by 9, 19, 49, or 99 fully-connected plain or highway layers and a single softmax output layer. The number of units in each layer is kept constant and it is 50 for highways and 71 for plain networks. That way the number of parameters is roughly the same for both.
To make the comparison fair we run a random search of 40 runs for both plain and highway networks to find good settings for the hyperparameters. We optimized the initial learning rate, momentum, learning rate decay rate, activation function for $H$ (either \textit{ReLU} or \textit{tanh}) and, for highway networks, the value for the transform gate bias (between -1 and -10). All other weights were initialized following the scheme introduced by \cite{He2015}. 

The convergence plots for the best performing networks for each depth can be seen in \autoref{fig:mnist-convergence}. While for 10 layers plain network show very good performance, their performance significantly degrades as depth increases. 
Highway networks on the other hand do not seem to suffer from an increase in depth at all. The final result of the 100 layer highway network is about 1 order of magnitude better than the 10 layer one, and is on par with the 10 layer plain network.
In fact, we started training a similar 900 layer highway network on CIFAR-100 which is only at 80 epochs as of now, but so far has shown no signs of optimization difficulties. 
It is also worth pointing out that the highway networks always converge significantly faster than the plain ones.

\subsection{Comparison to Fitnets}

\begin{table*}
	\centering
    \begin{tabular}{llll}
    \hline
    Network                                 & Number of Layers & Number of Parameters   & Accuracy \\ \hline
    Fitnet Results reported by \citet{Romero2014}  & ~                & ~                      & ~        \\
    \quad Teacher                                 & 5                & $\sim$9M    & 90.18\%  \\
    \quad Fitnet 1                                & 11               & $\sim$250K  & 89.01\%  \\
    \quad Fitnet 2                                & 11               & $\sim$862K  & 91.06\%  \\
    \quad Fitnet 3                                & 13               & $\sim$1.6M  & 91.10\%  \\
    \quad Fitnet 4                                & 19               & $\sim$2.5M  & 91.61\%  \\
    Highway networks                        & ~                & ~                      & ~        \\
    \quad Highway 1 (Fitnet 1)                    & 11               & $\sim$236K  & 89.18\%  \\
    \quad Highway 2 (Fitnet 4)                    & 19               & $\sim$2.3M  & \textbf{92.24\%}  \\
    \quad Highway 3*                               & 19               & $\sim$1.4M  & 90.68\%        \\
    \quad Highway 4*                               & 32               & $\sim$1.25M & 90.34\%        \\ \hline
    \end{tabular}
    \caption{CIFAR-10 test set accuracy of convolutional highway networks with rectified linear activation and sigmoid gates. For comparison, results reported by \citet{Romero2014} using maxout networks are also shown. Fitnets were trained using a two step training procedure using soft targets from the trained Teacher network, which was trained using backpropagation. We trained all highway networks directly using backpropagation. * indicates networks which were trained only on a set of 40K out of 50K examples in the training set.}
    \label{tab:fitnets}
\end{table*}

Deep highway networks are easy to optimize, but are they also beneficial for supervised learning where we are interested in generalization performance on a test set? To address this question, we compared highway networks to the thin and deep architectures termed \emph{Fitnets} proposed recently by \citet{Romero2014} on the CIFAR-10 dataset augmented with random translations. Results are summarized in \autoref{tab:fitnets}.

\citet{Romero2014} reported that training using plain backpropogation was only possible for maxout networks with depth up to 5 layers when number of parameters was limited to $\sim$250K and number of multiplications to $\sim$30M. Training of deeper networks was only possible through the use of a two-stage training procedure and addition of soft targets produced from a pre-trained shallow teacher network (hint-based training). Similarly it was only possible to train 19-layer networks with a budget of 2.5M parameters using hint-based training. 

We found that it was easy to train highway networks with number of parameters and operations comparable to fitnets directly using backpropagation. As shown in \autoref{tab:fitnets}, Highway 1 and Highway 4, which are based on the architecture of Fitnet 1 and Fitnet 4 respectively obtain similar or higher accuracy on the test set. We were also able to train thinner and deeper networks: a 19-layer highway network with $\sim$1.4M parameters and a 32-layer highway network with $\sim$1.25M parameter both perform similar to the teacher network of \citet{Romero2014}.

\section{Analysis}

\begin{figure*}[t]
\includegraphics[width=\textwidth]{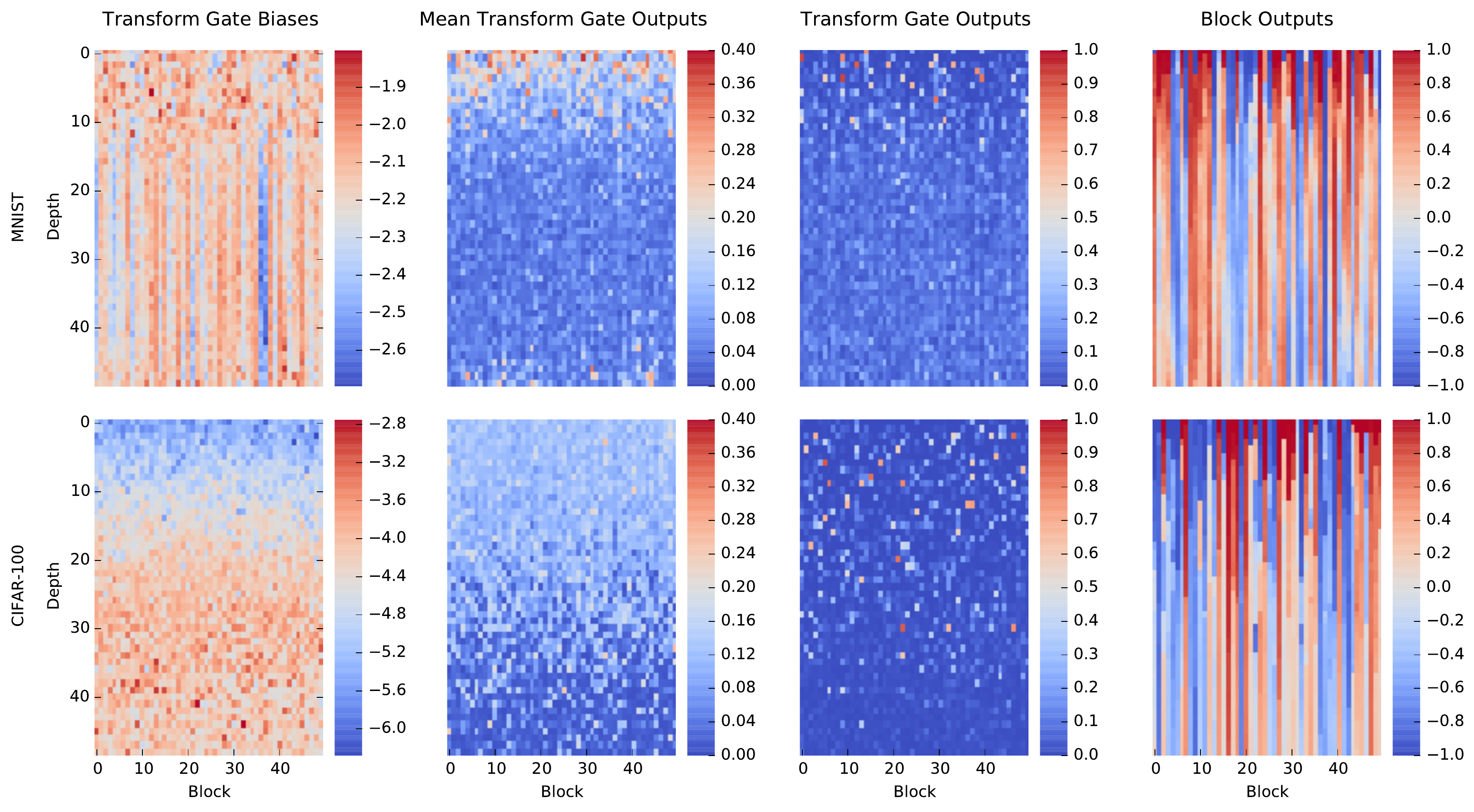}
\caption{Visualization of certain internals of the blocks in the best 50 hidden layer highway networks trained on MNIST (top row) and CIFAR-100 (bottom row). 
The first hidden layer is a plain layer which changes the dimensionality of the representation to 50. 
Each of the 49 highway layers (y-axis) consists of 50 blocks (x-axis).
The first column shows the transform gate biases, which were initialized to -2 and -4 respectively. 
In the second column the mean output of the transform gate over 10,000 training examples is depicted.
The third and forth columns show the output of the transform gates and the block outputs for a single random training sample. 
}
\label{fig:analysis}
\end{figure*}

In \autoref{fig:analysis} we show some inspections on the inner workings of the best\footnote{obtained via random search over hyperparameters to minimize the best training set error achieved using each configuration} 50 hidden layer fully-connected highway networks trained on MNIST (top row) and CIFAR-100 (bottom row). The first three columns show, for each transform gate, the bias, the mean activity over 10K random samples, and the activity for a single random sample respectively. The block outputs for the same single sample are displayed in the last column. 

The transform gate biases of the two networks were initialized to -2 and -4 respectively. 
It is interesting to note that contrary to our expectations most biases actually decreased further during training. 
For the CIFAR-100 network the biases increase with depth forming a gradient. 
Curiously this gradient is inversely correlated with the average activity of the transform gates as seen in the second column.
This indicates that the strong negative biases at low depths are not used to shut down the gates, but to make them more selective. 
This behavior is also suggested by the fact that the transform gate activity for a single example (column 3) is very sparse. 
This effect is more pronounced for the CIFAR-100 network, but can also be observed to a lesser extent in the MNIST network.

The last column of \autoref{fig:analysis} displays the block outputs and clearly visualizes the concept of ``information highways''.
Most of the outputs stay constant over many layers forming a pattern of stripes. 
Most of the change in outputs happens in the early layers ($\approx 10$ for MNIST and $\approx 30$ for CIFAR-100). 
We hypothesize that this difference is due to the higher complexity of the CIFAR-100 dataset. 

In summary it is clear that highway networks actually utilize the gating mechanism to pass information almost unchanged through many layers. 
This mechanism serves not just as a means for easier training, but is also heavily used to route information in a trained network.
We observe very selective activity of the transform gates, varying strongly in reaction to the current input patterns.

\section{Conclusion}

Learning to route information through neural networks has helped to scale up their application to challenging problems by improving credit assignment and making training easier \cite{Srivastava2014}. Even so, training very deep networks has remained difficult, especially without considerably increasing total network size.

Highway networks are novel neural network architectures which enable the training of extremely deep networks using simple SGD. While the traditional plain neural architectures become increasingly difficult to train with increasing network depth (even with variance-preserving initialization), our experiments show that optimization of highway networks is not hampered even as network depth increases to a hundred layers.

The ability to train extremely deep networks opens up the possibility of studying the impact of depth on complex problems without restrictions. Various activation functions which may be more suitable for particular problems but for which robust initialization schemes are unavailable can be used in deep highway networks. Future work will also attempt to improve the understanding of learning in highway networks.

\section*{Acknowledgments}
This research was supported by the by EU project ``NASCENCE'' (FP7-ICT-317662).
We gratefully acknowledge the support of NVIDIA Corporation with the donation of the Tesla K40 GPUs used for this research.

\bibliography{highways}
\bibliographystyle{icml2015}
\end{document}